\documentclass{article}

\PassOptionsToPackage{numbers, compress}{natbib}

 \usepackage[preprint]{nips_2018}

\usepackage[T1]{fontenc}    
\usepackage{url}            
\usepackage{booktabs}       
\usepackage{amsfonts}       
\usepackage{nicefrac}       
\usepackage{microtype}      
\usepackage{tabularx}
\usepackage{algorithm}
\usepackage{algorithmic}
\usepackage{epsfig}
\usepackage{amsmath}
\usepackage{multirow}
\usepackage{subfigure}

\title{Graph Laplacian Regularized Graph Convolutional Networks for Semi-supervised Learning}

\author{
  Bo Jiang, Doudou Lin\\
  School of Computer Science and Technology \\
  Anhui University\\
  Hefei, China \\
  \texttt{jiangbo@ahu.edu.cn} \\
}

\begin{document}

\maketitle

\begin{abstract}
Recently, graph convolutional network (GCN) has been widely used for semi-supervised classification and deep feature representation on graph-structured data.
However, existing GCN generally fails to consider the local invariance constraint in learning and representation process.
That is,
if two data points $X_i$ and $X_j$ are close in the intrinsic geometry of the
data distribution, then their labels/representations should also be close to each other.
This is known as \emph{local invariance assumption} which plays an essential role in the development of various kinds of traditional algorithms, such as dimensionality reduction and semi-supervised learning, in machine learning area.
To overcome this limitation, we introduce a graph Laplacian GCN (gLGCN) approach for graph data representation and
semi-supervised classification.
The proposed gLGCN model is capable of encoding both graph structure and node features together while maintains the local invariance constraint naturally for robust data representation and  semi-supervised classification.
Experiments show the benefit of the benefits the proposed gLGCN network.

\end{abstract}

\section{Introduction}
Given a graph $G(V, E)$ with $V$ denoting the $n$ nodes and $E$ representing the edges.
Let $A \in \mathbb{R}^{n\times n}$ be the corresponding adjacency matrix, and $X=(X_1,X_2,\cdots X_n)\in \mathbb{R}^{p\times n}$ be the collection of node features where $X_i$ denotes the feature descriptor for node $v_i\in V$.
For semi-supervised learning tasks, let ${L}$ indicates the set of labelled nodes and
$Y_{{L}}$ be the corresponding labels for labelled nodes.
The aim of semi-supervised learning is to predict the labels for the unlabelled nodes.

\subsection{Graph Laplacian regularization}

One kind of popular method for semi-supervised learning problem is to
use graph-based semi-supervised learning, where  the label information is smoothed over the graph via
graph Laplacian regularization~\cite{belkin2006manifold,zhu2003semi} i.e.,
\begin{equation}\label{EQ:GLR}
\mathcal{L} = \mathcal{L}_{\mathrm{label}} + \lambda \mathcal{L}_{\mathrm{reg}}
\end{equation}
Here $ \mathcal{L}_{\mathrm{label}}$ and $\mathcal{L}_{\mathrm{reg}}$ are defined as,
\begin{equation}\label{EQ:fit}
\mathcal{L}_{\mathrm{label}} = \sum\nolimits_{i\in L} l(Y_i, f(X_i)) \ \  \ \ \ \mathrm{and} \ \ \ \ \mathcal{L}_{\mathrm{reg}} = \sum^n\nolimits_{i,j=1}S_{ij} \|f(X_i) - f(X_j)\|^2 
\end{equation}
where $l(\cdot)$ denotes some standard supervised loss function  and $\mathcal{L}_{\mathrm{reg}}$ is called
as graph Laplacian regularization. Function $f(X_i)$ denotes the label prediction of node $v_i$ and $S_{ij}$ denotes some kind of relationship (e.g., affinity and similarity) between graph node $v_i$ and $v_i$.
We can set $S$ as adjacency matrix $A$ or some other graph construction.
One traditional graph is to use a k nearest neighborhood graph with edge weighted by some kernel metric (e.g., Gaussian kernel) $K(X_i, X_j)$ between feature $X_i$ and $X_j$.
Parameter $\lambda >0$ balances two terms.
The objective function Eq.(1) encourages that if two data points $X_i$ and $X_j$ are close in data distribution, then their corresponding labels should also be close with each other.


\subsection{Graph convolutional network}

Recently, graph convolutional network (GCN) ~\cite{defferrard2016convolutional,kipf2016semi} has been proposed for semi-supervised tasks.
It aims to seek a nonlinear function $f(X,A)$ to predict labels for unlabelled nodes.
It contains several propagation layers and one final perceptron layer together. Given any input feature $X$ and graph structure (adjacency matrix) $A$, GCN conducts
the following layer-wise propagation rule ~\cite{kipf2016semi}, 
\begin{align}
& X^{(1)}  = \mathrm{ReLu}(\widetilde{A}XW^{(0)}) \nonumber \\
& \,\,\,\,\,\,\,\, \cdots \nonumber \\
& X^{(K)} = \mathrm{ReLu}(\widetilde{A}X^{(K-1)}W^{(K-1)})  \\
&Z = \mathrm{softmax} (\widetilde{A}X^{(K)} W^{(K)} ) \nonumber
\end{align}
%
Here, $\widetilde{A}=\bar{D}^{-1/2}\bar{A}\bar{D}^{-1/2}$ and
$\bar{A}=A +I$, where $I$ is the  identity matrix and $\bar{D}$ is a diagonal matrix with
$\bar{D}_{ii}=\sum_j\bar{A}_{ij}$.
$\{X^{(1)}, X^{(2)},\cdots X^{(K)}\}$ denotes the feature output of the different layers and $Z$ is the label output of the final layer where $Z_i$ is the label indication vector of the node $v_i$. For semi-supervised learning, the optimal weights $\{W^{(0)}, W^{(1)},\cdots W^{(K)}\}$ can be trained  by minimizing the following cross-entropy loss function over all labeled nodes $L$.
 \begin{equation}
\mathcal{L}_{\mathrm{GCN}} = -\sum\nolimits_{i\in L} \sum^d\nolimits_{j=1} Y_{ij}\mathrm{ln} Z_{ij}
 \end{equation}
\noindent \textbf{Remark.}
When $W^{(k)}\in \mathbb{R}^{d_{k-1}\times d_k}$ and $d_{k-1}<d_k$, the above GCN provides a series of low-dimensional embedding $X^{(k)}$ for the original input feature $X$. 



\section{Graph Laplacian GCN}

In this section, we present two types of graph Laplacian GCN.
Inspired by traditional graph based semi-supervised learning model, we first propose a graph Laplacian
GCN for robust semi-supervised learning.
In addition, motivated by manifold assumption, we propose to incorporate manifold regularization in GCN feature representation.

\subsection{Graph Laplacian label prediction}

GCN ~\cite{kipf2016semi}  predicts the labels for unlabelled nodes by using label propagation on graph.
One limitation of GCN is that it  fails to consider the local consistency of nodes with similar features in label propagation, i.e., if the features of neighboring node $v_i$ and $v_j$ are similar, then their corresponding labels should also be close.
This point has commonly used in traditional graph based semi-supervised learning model \cite{belkin2006manifold,weston2012deep}.
This motivate us to propose an improved graph Laplacian GCN (gLGCN), which aims to conduct local label propagation via GCN while maintains the local consistency via graph Laplacian regularization. This can be obtained by optimizing the following loss function,
 \begin{align}
&\mathcal{L}_{\mathrm{gLGCN}}(Z) = \mathcal{L}_{\mathrm{GCN}}(Z) + \lambda \mathcal{L}_{\mathrm{reg}}(Z)\nonumber\\
&= -\sum\nolimits_{i\in L} \sum^d\nolimits_{k=1} Y_{ik}\mathrm{ln} Z_{ik} + \lambda \sum^n\nolimits_{i,j=1}S_{ij}\|Z_i - Z_j\|^2
 \end{align}
where $Z_i$ denotes the $i$-th row of matrix $Z$ and $S_{ij}$ denotes the similarity between node $i$ and $j$, as mentioned in Eq.(2).
\begin{figure}
\centering
\centering
 \includegraphics[width=0.99\textwidth]{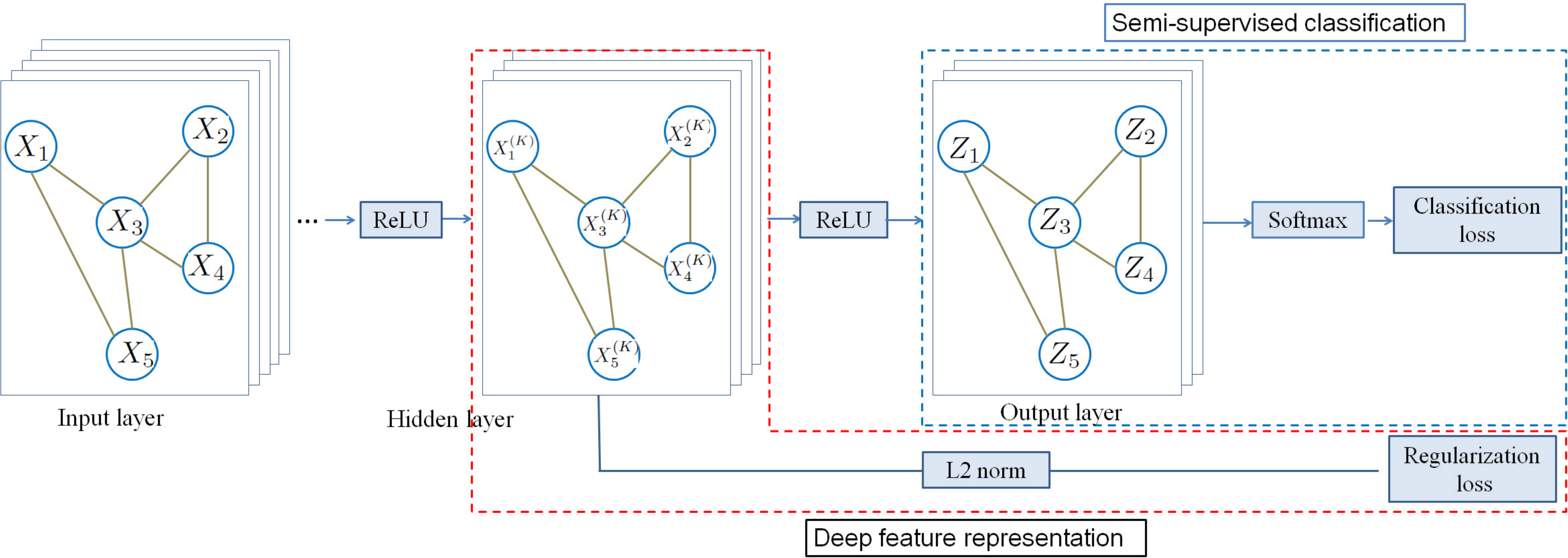}
  \caption{ Architecture of the proposed Graph Laplacian GCN for robust data representation. }\label{fig::lambda}
\end{figure}
\subsection{Graph Laplacian feature representation}

On the other hand,
to boost the effectiveness of the learned deep representation, similar to \cite{weston2012deep,te2018rgcnn}, we also incorporate the
graph Laplacian regularization into the feature generation layer and introduce a
regularization loss, as shown in Figure 1.

The regularization subnetwork can be regarded as a kind of pair-wise siamese network,
 which first takes two low-dimensional features of the final layer $X^{(K)}_i$  and $X^{(K)}_j$  as input,
and then calculates the distance between them.
If the similarity $S_{ij}$ between node $v_i$ and $v_j$ is larger, then the generated low-dimensional feature representation
$X^{(K)}_i$  and $X^{(K)}_j$ should be close with each other.
This is known as manifold assumption which has been widely used in dimensionality reduction in machine learning area.
We adopt a loss function as
 \begin{align}
\mathcal{L}_{\mathrm{reg}}(X^{(K)}) =  \sum^n\nolimits_{i,j=1}S_{ij}\|X^{(K)}_i - X^{(K)}_j\|^2
 \end{align}
The above semi-supervised learning and manifold regularized feature representation
are optimized at the same time
in a unified network. Thus, we can write the total objective function as
\begin{align}
\mathcal{L} = \mathcal{L}_{\mathrm{GCN}}(Z) + \lambda \mathcal{L}_{\mathrm{reg}}(X^{(l)})
\end{align}
where $\lambda >0$ is the balanced parameter.

\noindent \textbf{Comparison with related works.}
Our model is different from previous works \cite{te2018rgcnn,weston2012deep} in several aspects.
First, we focus on semi-supervised learning problem. The proposed regularization on the final layer provides a label propagation for semi-supervised learning problem. 
Second, in our model, the feature $X^{(K)}_i$ provides a low-dimensional representation for graph node $v_i$ and thus
  our model provides a kind of local preserving low-dimensional embedding for semi-supervised learning. 
Third,  the proposed model conducts feature propagation on graph $\widetilde{A}$ and linear projection via $W^{(k)}$ together in each layer of the network. In contrast, in previous work \cite{weston2012deep}, it only conducts linear projection in each layer.
Overall, it integrates the benefits of work \cite{weston2012deep} and GCN ~\cite{kipf2016semi} simultaneously for semi-supervised learning.

In addition, for semi-supervised learning, the labels of some nodes are known. We can  define the label correlation $C_{ij}$ as follows,
$$
C_{ij} =
\begin{cases}
1 & \text{if} \  \ v_i, v_j \in L \ \ \text{and} \ \ Y_i=Y_j\\
-\alpha & \text{if} \  \ v_i, v_j \in L \ \ \text{and} \ \ Y_i\neq Y_j\\
0 & \text{otherwise}
\end{cases}
$$
Based on $C$, we can incorporate the label information via the regularization loss as 
 \begin{align}
\mathcal{L}_{\mathrm{reg}}(X^{(K)}) =  \sum^n\nolimits_{i,j=1}C_{ij}\|X^{(K)}_i - X^{(K)}_j\|^2
 \end{align}
which is similar to the widely used triplet loss function used in deep networks.
%
\section{Evaluation}

To evaluate the effectiveness of the proposed gLGCN network.
We follow the experimental setup in
work ~\cite{Yang:2016} and test our model on  the citation network datasets including Citeseer, Cora and Pubmed~\cite{sen2008collective}
The detail introduction of datasets used in
our experiments are summarized in Table 1.
The optimal regularization parameter $\lambda$ is chosen based on
validation.
\begin{table}[!htp]
\centering
\caption{Dataset description in experiments}
\centering
\begin{tabular}{c|c|c|c|c|c|c}
  \hline
    \hline
  Dataset & Type & Nodes & Edges & Classes & Features & Label rate \\
  \hline
  Citeseer & Citation network & 3327 & 4732 & 6 & 3703 & 0.036 \\
  Cora & Citation network & 2708 & 5429 & 7 & 1433 & 0.052 \\
  Pubmed & Citation network & 19717 & 44338 & 3 & 500 & 0.003 \\
  \hline
    \hline
\end{tabular}
\end{table}

We compare against the same baseline methods including traditional label propagation
(LP)~\cite{zhu2003semi}, semi-supervised embedding (SemiEmb) ~\cite{weston2012deep}, manifold
regularization (ManiReg)~\cite{belkin2006manifold}, Planetoid ~\cite{Yang:2016} and graph convolutional network (GCN)~\cite{kipf2016semi}.
For GCN~\cite{kipf2016semi}, we implement it using the pythorch code provided by the authors. 
For fair comparison, we also implement our gLGCN by using pythorch.
Results for the other baseline methods are taken from work ~\cite{Yang:2016,kipf2016semi}
For component analysis, we implement it with three versions, i.e.,
1) gLGCN-F that incorporates Laplacian regularization in feature representation. 
2) gLGCN-L that incorporates Laplacian regularization in label prediction.
3) gLGCN-F-L that incorporates Laplacian regularization in both feature representation and label prediction.
Table 2 summarizes the comparison results. Here we can note that, our gLGCN performs better than traditional LP, ManiReg and GCN, which
clearly indicates the benefit of the proposed gLGCN network method.

\begin{table}[!htp]
\centering
\caption{Comparison results on different datasets}
\centering
\begin{tabular}{c|c|c|c}
  \hline
    \hline
  Methond & Citeseer & Cora & Pubmed  \\
  \hline
  ManiReg\cite{belkin2006manifold}  & 60.1 & 59.5 & 70.7  \\
  SemiEmb\cite{weston2012deep} & 59.6 & 59.0 & 71.1  \\
  LP\cite{zhu2003semi} & 45.3 & 68.0 & 63.0  \\
  Planetoid\cite{Yang:2016}& 64.7 & 75.7 & 77.2  \\
  GCN\cite{kipf2016semi} & 70.4 & 81.4 & 78.6  \\
  gLGCN-F & 70.8 & 82.2 & 79.2  \\
  gLGCN-L & 71.3 & 82.7 & 79.2  \\
  gLGCN-F-L & 71.4 & 83.3 & 79.3  \\
  \hline
    \hline
\end{tabular}
\end{table}

\bibliographystyle{ieee}
\bibliography{nmfgm}

\end{document}